# Co-supervised learning paradigm with conditional generative adversarial networks for sample-efficient classification

Hao Zhen[1], Yucheng Shi[2], Jidong J. Yang[1] and Javad Mohammadpour Vehni[3]

[1]Smart Mobility and Infrastructure Laboratory, College of Engineering,
University of Georgia, Athens GA, USA

[2]Department of Computer Science, University of Georgia, Athens GA, USA

[3]Department of Mechanical Engineering, Clemson University; Clemson SC, USA

## Abstract

Classification using supervised learning requires annotating a large amount of classes-balanced data for model training and testing. This has practically limited the scope of applications with supervised learning, in particular deep learning. To address the issues associated with limited and imbalanced data, this paper introduces a sample-efficient co-supervised learning paradigm (SEC-CGAN), in which a conditional generative adversarial network (CGAN) is trained alongside the classifier and supplements semantics-conditioned, confidence-aware synthesized examples to the annotated data during the training process. In this setting, the CGAN not only serves as a *co-supervisor* but also provides complementary quality examples to aid the classifier training in an end-to-end fashion. Experiments demonstrate that the proposed SEC-CGAN outperforms the external classifier GAN (EC-GAN) and a baseline ResNet-18 classifier. For the comparison, all classifiers in above methods adopt the ResNet-18 architecture as the backbone. Particularly, for the Street View House Numbers dataset, using the 5% of training data, a test accuracy of 90.26% is achieved by SEC-CGAN as opposed to 88.59% by EC-GAN and 87.17% by the baseline classifier; for the highway image dataset, using the 10% of training data, a test accuracy of 98.27% is achieved by SEC-CGAN, compared to 97.84% by EC-GAN and 95.52% by the baseline classifier.

## 1. Introduction

Modern machine learning methods require a large amount of data for training and testing. In supervised learning methods, data annotation is laborious and time-consuming and sometimes requires professional judgement. This has practically limited the scope of applications with supervised learning algorithms, in particular deep learning. In some cases, the overall quantity of data is adequate, but class distribution of the data is imbalanced, where minority classes are rarely observed. For example, in the case of classifying weather from highway images, some weather conditions (e.g., foggy and snowy) are rare than others (e.g., sunny and cloudy). However, these rare cases often lead to dangerous driving situations for motorists, and thus demand a high classification accuracy in the context of road weather alert. This nature-inherited data imbalance imposes a learning bias, which is undesirable for classifying rare events of severe consequences. This paper aims to improve supervised classification tasks for either a small size and/or imbalanced datasets constrained by class distribution.

For small and imbalanced datasets, researchers usually resort to data augmentation techniques to attain better performance. For image data augmentation, the traditional method is label-preserving transformations [1], such as rotation, flipping, cropping, and reflection; another method is using synthetic data. For the latter, GANs have been used for augmentation in data scarce domains, such as medical images [2], where synthetic data is generated and utilized to augment small datasets for training classification models, leading to improved classification results. This is due to the fact that the traditional augmentation methods do not contribute novel features rather sample from a low-level, high-dimensional feature space, while GAN-generated images contain distinctive high-level features sampled from a low-dimensional feature space or manifold. However, professional knowledge may still be needed to manually screen qualified synthetic data for classification. The screening process can be laborious. For synthetic images selection, different sampling methods based on class conditional probability, realism conditional probability, or reinforcement learning were studied in [3], which showed that class conditional probability produced by a pretrained classifier outperformed the realism condition probability and RL-based sampling methods. More recently, Conditional GANs (CGANs) [4] have been used for image augmentation. Synthetic images generated by pretrained CGAN were utilized to augment real data, resulting in improved classification accuracy for fruit quality classification [5]. In addition, research [6] demonstrated that deep conditional generative models, including deep CGANs, have the ability to balance the imbalanced data, which lead to better classification performance. However, these GAN-based methods rely on synthetic images from pretrained GAN models, and the training is not end-to-end.

There are some existing end-to-end, semi-supervised shared architecture network for classification [7–9]. But sharing a base network by the discriminator and the classifier can be too restrictive and likely degrade performance of both. Some efforts have been attempted to disentangle the discriminator and classifier networks, such as Triple-GAN [10] and EC-GAN [11]. Nonetheless, Triple-GAN focuses more on the discrimination task and does not use synthesized data for augmentation for the classification. On the other hand, EC-GAN tackles low-sample classification task by combining GANs and semi-supervised learning but does not have control over the semantics (labels) of generated images and the sampling distribution. Thus, it does not address the data imbalance issue. With the popularity of unsupervised and self-supervised learning methods in recent years [12–14], supervised learning can benefit from using backbones pretrained with these *label-free* methods. Nevertheless, this does not solve the data imbalance problem.

In summary, we propose a simple end-to-end paradigm that deals with small and imbalanced datasets for supervised classification. The goal is to properly augment small and imbalanced datasets with quality synthesized data to improve supervised classification.

Synthesized samples are commonly produced by generative models that learn the underlying distribution of data. Image synthesis models provide a unique opportunity for performing semi-supervised learning as these models build a rich prior over natural image statistics that can be leveraged by classifiers to improve predictions on datasets for which few labels exist [15]. Generative adversarial networks (GANs) [16] is one of the most popular generative modeling frameworks. Since its inception, many variants of the GAN have been proposed to improve the quality of generated images [17–22].

We adopt Conditional GANs (CGANs) [4] in the proposed SEC-CGAN paradigm to generate synthesized samples with underlying labels that resemble real data. It should be noted that the proposed paradigm is generic and can be extended to other types of conditional GANs, such as AC-GANs [23] and IcGANs [24].

Particularly, we leverage the image-synthesizing capacity of CGANs to address the limitations of small or imbalanced datasets for supervised classification tasks. A new learning framework is introduced, which consists of a sample-efficient classifier (SEC) learned in a fully supervised fashion from both annotated data and synthesized data generated by a CGAN that is trained alongside the classifier. We refer to this method as SEC-CGAN and the learning scheme from both annotated data and synthesized data as *co-supervised* learning in this paper. In this setting, the classes (labels) of generated images can be controlled, leading to efficient sampling of minority classes. The generated images come with labels as well as associated confidences. The latter is used to ensure that the quality synthesized samples are used for the classifier training, which is similar to the realism conditional probability sampling in [3]. The key difference lies in that the GAN in [3] is pretrained while the CGAN in the proposed SEC-CGAN method is trained together with the classifier in an end-to-end fashion. Our experiments demonstrate the superiority of the SEC-CGAN over the state-of-the-art small-sample method, EC-GAN [11], and a baseline classifier.

The remainder of this paper is organized as follows. The related works are first reviewed and discussed in Section 2. The proposed SEC-CGAN is introduced in Section 3, followed by the experiments in Section 4. Conclusions are presented in Section 5.

## 2. Related Works

In this section, the existing work pertaining to the scope of this study is discussed. We first introduce the background of GANs, followed by CGANs. Then, pseudo-labeling and shared architecture, which are current existing end-to-end works that leverage GANs for better classification or other tasks, are discussed subsequently.

### 2.1. GANs

GANs are designed to generate images by training two neural networks: a generator $G$ and a discriminator $D$. In this framework, the generator tries to generate samples approximating the implicit distribution of real data while the discriminator tries to differentiate the generated data from the real data by predicting the likelihood that a generated sample is from the real data. This can be viewed as two networks playing a minmax game. During the training process, the generator is updated to produce better images to fool discriminator, while the discriminator improves at differentiating images as real or fake. The minmax objective function is defined as:

$$\min_{G} \max_{D} V(D,G) = E_x\big[log\big(D(x)\big)\big] + E_z\Big[log\Big(1 - D\big(G(z)\big)\Big)\Big] \tag{1}$$

where $E_x$ is the expectation over the data, $log[D(x)]$ is the discriminator's predicted log probability of an image being real, and $E_z$ is the expectation over the noise prior $p_z$. Theoretically, the loss is minimized with a value of -log4 when the distribution of $G(z)$ converges to the distribution of the data [16], signaling the equilibrium is reached, at which point the generator is creating perfectly realistic images and the discriminator has a 50% chance to differentiate the generated one from real one. For the proposed method, the deep convolutional GAN (DCGAN) architecture [17] is adopted, which is a commonly used GAN architecture for image synthesis. In DCGAN, the generator sequentially up-samples the input features using fractionally-strided convolutions, whereas the discriminator uses regular convolution to classify the input images.

## 2.2. CGANs

The CGANs [4] are a special class of GANs that incorporate a conditional setting for generator and discriminator based on some prior attributes. Examples of conditions are class labels [15, 23, 24], text [25, 26], or images including image translation [22, 27] and style transfer [28]. Most CGANs apply auxiliary conditional information on both generator and discriminator networks by concatenating it to the inputs. Apart from that, the condition information had been included in the batch normalization [28]. Similar to GANs, CGANs can also be viewed as a two-player game with the following value function:

$$\min_{G} \max_{D} V(D,G) = E_{x \propto p_{data}(x)}[logD(x|y)] + E_{z \propto p_z(z)}\Big[log\Big(1 - D\big(G(z|y)\big)\Big)\Big] \tag{2}$$

where $D\,(x\,|\,y)$ and G $(z\,|\,y)$ are the discriminator and generator functions for given label $y$, respectively. $V$ is the value function. $E$ denotes expectation.

One desirable feature of CGANs is that the generated images already have auxiliary conditional information (i.e., labels), which can be leveraged in supervised learning for classification tasks. Patel et al. [29] proposed data augmentation with CGANs for automatic modulation classification and indicated that CGANs can benefit classification with limited data. In the proposed SEC-CGAN, a conditional DCGAN is utilized to generate synthesized images, which are then passed along with their labels for supervised classification tasks. We set a confidence threshold for the discriminator to decide on the quality of synthesized images for use in classifier training since using poorly generated images would hurt the learning of the classifier.

## 2.3. Pseudo-Labeling

Another data augmentation approach is pseudo-labeling. Pseudo-labeling aims to produce artificial labels for unlabeled images [30]. In most cases, it pseudo-labels the unlabeled data by learning from the labeled dataset. The model simply chooses the class that has the highest predicted probability according to its current state as the 'label' for unlabeled data. Pseudo-labeling is similar to entropy regularization [31].

Recently, Haque [11] applied pseudo-labeling for the fake, unlabeled images generated by the GAN models for classification; the method was named EC-GAN. However, the EC-GAN does not have control over the semantics (labels) of the generated data and likewise the class distribution of the data. This may cause a bias due to the unbalanced data sampling. In contrast with pseudo-labeling, our proposed SEC-CGAN controls which classes to generate as well as the quality of the synthesized images, leading to efficient sampling of minority classes and high quality of synthesized images.

### 2.4. Shared architecture

Many existing methods using GANs for semi-supervised learning adopted a single network (see Figure 1) for classifier and discriminator [7–9]. There are two main mechanisms to combine GAN with a classifier network: one is to consider two different final layers coming from the same network for discriminator and classifier, and the other is to consider two separate head networks from the same base network with one for discriminating between 'real' and 'fake' (discriminator) and one for predicting class labels (classifier). However, sharing a base network for two different tasks, i.e., discrimination and classification, may degrade performance for both tasks [11]. The loss function for the shared discriminator model is expressed as:

$$BCE\left(D_d\left(D(G(z))\right), 0\right) + BCE\left(D_d(D(x)), 1\right) + CE\left(D_c(D(x)), y\right) \qquad (3)$$

where $x$ is the real data and $y$ is the corresponding label, $z$ is a randomly generated vector from a prior distribution, $BCE$ denotes binary cross-entropy, $CE$ denotes cross-entropy, $D_d$ denotes the discriminator head, $D_c$ denotes the classifier head, $D$ denotes the discriminator backbone, and $G$ is the generator. In this shared architecture, the discriminator head and classifier head each independently update the shared base network parameters along with their own individual final layers. This means that the network attempts to minimize two separate losses with the shared backbone parameters, which is the primary concern for the performance degradation.

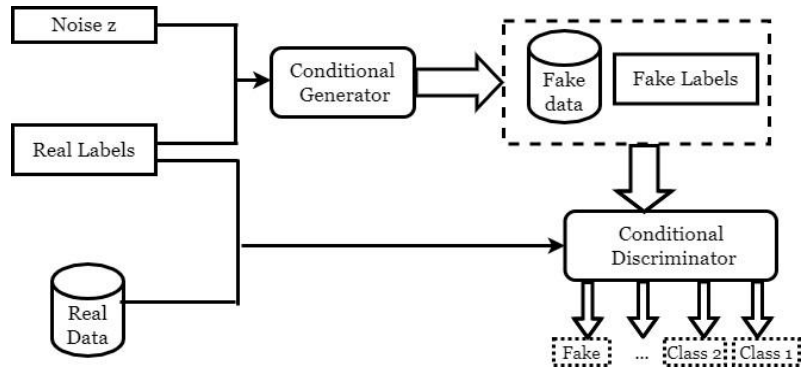

**Figure 1.** Shared discriminator architecture.

There have been some other efforts on disentangling the discriminator and classifier by allowing the discriminator to have its sole task for discrimination. For example, the Triple GAN method in [10] applies a separate classifier network for semi-supervised learning using GANs, resulting in three different sub-networks: a generator, a discriminator, and a classifier. The generator generates synthesized image-label pairs while the classifier uses pseudo-labelling to label the unlabeled images in the training dataset. The discriminator only has the role of identifying fake image-label pairs.

Similar to the Triple GAN, the proposed SEC-CGAN uses separate discriminator and classifier. Instead of pseudo-labeling, our method employs a CGAN as a co-supervisor that controls the class distribution and quality of generated samples to enhance the classifier training.

## 3. SEC-CGAN

The SEC-CGAN consists of three networks: a conditional generator, a conditional discriminator, and a classifier as Figure 2 depicts, which is different from the traditional shared architecture, where the discriminator and classifier shared the same deep convolution architecture and weights [7–9].

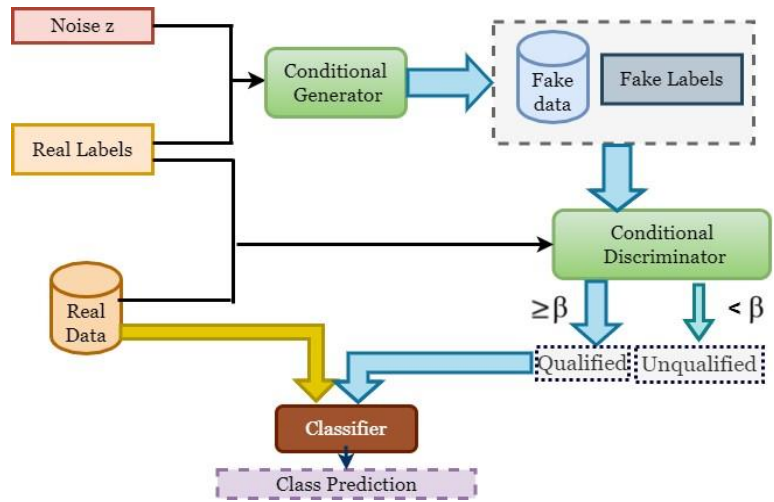

**Figure 2.** The proposed SEC-CGAN paradigm.

### 3.1. Model architecture

The conditional generator, the conditional discriminator, and the classifier each has its independent architecture as depicted in Figure 2. A random vector and a label are fed to the conditional generator to generate an image. We use the conditional generator in the conditional DCGAN architecture [4], which has an embedding layer, convolutional-transpose layers, batch normalization layers, and ReLU activations. For the conditional discriminator, we use an architecture similar to the generator, but substitute convolutional-transpose layers with strided convolutional layers and use leakyReLU activations. The discriminator uses one-hot vector of the label to indicate the condition for the discriminator output. For the classifier network, we use ResNet-18, a classic image classification model with proven empirical success [32]. For the classifier training, both the real data and synthesized samples are utilized. Specifically, the generator supplements classifier training with synthesized samples along with labels. The quality of synthesized samples is controlled by a confidence threshold of the discriminator. This ensures that only the qualified synthesized images are utilized.

## 3.2. Training

At every training iteration, the conditional generator is given random vectors and labels for producing synthesized images. The discriminator is then employed to distinguish between real and generated samples. Simultaneously, a classifier is trained in a supervised fashion not only on available real data-label pairs, but also on qualified synthesized images and their conditional labels.

To decide whether a synthesized sample is qualified or containing useful information for classifier training, the synthesized images and labels are fed to the discriminator and a confidence threshold is used to determine the desired quality of the synthesized samples. This means if the probability of a synthesized sample to be real is larger than the confidence threshold, the sample is qualified for use in training of the classifier. Otherwise, it will be discarded. The role of the discriminator is not only to differentiate the fake samples from real ones, but also to serve as a 'guardian' to select high-confidence synthesized samples to enhance the classifier training. In the training setting, the newly synthesized images are produced for each minibatch and immediately fed to the classifier training if their confidences exceed the predetermined threshold. Weighted batch sampling is adopted to balance the class distribution within each batch during training to mitigate the data imbalance problem. For every minibatch, the generated synthesized data is balanced before feeding to the discriminator to decide whether it can be used for classifier training. It should be noted that the weighted batch sampling can be adapted with other suitable sampling methods for different tasks, such as [33]. For the SEC-CGAN, the discriminator loss is defined as:

$$L_D(x, y, z) = BCE[D(x, y), 1] + BCE[D(G(z, \hat{y}), \hat{y}), 0]$$
(4)

where $BCE$ denotes binary cross-entropy, $D$ is the discriminator, $G$ is the generator, $x$ is real input, $y$ is the real label. $z$ is a random vector and $\hat{y}$ is the label for synthesized image, $G(z, \hat{y})$. The first term is the loss of the discriminator on real images. The second term is the loss of the discriminator on synthesized images produced by the generator. The generator loss and classifier loss are computed by Eq. (5) and Eq. (6), respectively.

$$L_G(x, y, z) = BCE[D(G(z, \hat{y}), \hat{y}), 1]$$
(5)

$$L_C(x, y, z, \hat{y}) = CE[C(x), y] + \lambda \cdot CE[C(G(z, \hat{y})), \hat{y} \mid D(G(z, \hat{y}), \hat{y}) \geq \beta]$$
(6)

where $\lambda$ is a weighting multiplier that controls the relative contribution of generated data to the classifier training and $\beta$ is the confidence threshold that controls the quality of data to be used for training. In Eq. (6), the first term is the loss computed from the training of the classifier on the real data. The second term is the loss computed from the training of the classifier on the synthesized data selected by the discriminator with a confidence being greater than or equal to $\beta$. Algorithm 1 shows how the SEC-CGAN can be trained by stochastic gradient descent. In the experiment, we used Adam optimizer [35]. The code is available at https://github.com/SMIL-AI/SEC-CGAN.

---

**Algorithm 1** SEC-CGAN training

---

**Input:** Minibatch, which includes $m$ real sample pairs $(x, y)$ and class-balanced $k$ noise and label pairs $(z, \hat{y})$.

**for** number of iterations **do**

    Update the parameters for discriminator $D$ on real data $(x, y)$.

$$\theta_d \leftarrow \theta_d + \eta \nabla_{\theta_d} \frac{1}{m} \sum_{i=1}^{m} log\left(D(x^{(i)}, y^{(i)})\right) \qquad \text{per the first loss term in Eq. (4)}$$

    Update the parameters for discriminator $D$ on generated data $(G(z, \hat{y}), \hat{y})$.

$$\theta_d \leftarrow \theta_d + \eta \nabla_{\theta_d} \frac{1}{k} \sum_{i=1}^{k} log\left(1 - D(G(z^{(i)}, \hat{y}^{(i)}), \hat{y}^{(i)})\right) \quad \text{per the second loss term in Eq. (4)}$$

    Update the parameters for generator $G$ on generated data $(G(z, \hat{y}), \hat{y})$.

$$\theta_g \leftarrow \theta_g + \eta \nabla_{\theta_g} \frac{1}{k} \sum_{i=1}^{k} log\left(D(G(z^{(i)}, \hat{y}^{(i)}), \hat{y}^{(i)})\right) \qquad \text{per Eq. (5)}$$

    Update the parameters for classifier $C$ on real data $(x, y)$.

$$\theta_c \leftarrow \theta_c + \eta \nabla_{\theta_c} \frac{1}{m} \sum_{i=1}^{m} y^{(i)} log\left(C(x^{(i)})\right) \qquad \text{per the first loss term in Eq. (6)}$$

    Collect $k'$ generated samples that satisfy $D(G(z, \hat{y}), \hat{y}) \geq \beta$.

    Update the parameters for classifier $C$ on the qualified $k'$ generated sample pairs.

$$\theta_c \leftarrow \theta_c + \lambda \eta \nabla_{\theta_c} \frac{1}{k'} \sum_{i=1}^{k'} \hat{y}^{(i)} log\left(C(G(z^{(i)}, \hat{y}^{(i)}))\right) \quad \text{per the second loss term in Eq. (6)}$$

**end for**

---

## 4. Experiments

To demonstrate the effectiveness of the proposed SEC-CGAN, we evaluated it against the EC-GAN and a baseline classifier using two datasets. The first one is the public Street View House Numbers (SVHN) dataset [34] and the second one is a road weather dataset collected by our team in Smart Mobility and Infrastructure Lab at the University of Georgia. We used Adam optimizer [35] for training.

### 4.1. SVHN dataset

To test SEC-CGAN's performance on small sample datasets, we define 5 scenarios of incremental training datasets, including randomly sampled 5%, 10%, 15%, 20%, 25% of the original training dataset. The models are evaluated using the full test dataset, containing 26,032 images. The learning rate for the generator, the discriminator, and the classifier are all set to 2e-4. The multiplier $\lambda$ and threshold $\beta$ are set to 0.6 and 0.7, respectively. For comparison, we also evaluated two additional models: EC-GAN [11] and a baseline ResNet-18 classifier [32]. The EC-GAN is the state-of-the-art semi-supervised method for small-sample classification, which uses pseudo labeling method to connect the GAN with an external classifier. All classifiers adopt the ResNet-18 architecture. The baseline ResNet-18 classifier is trained only with the annotated data.

All models are trained with label-preserving data augmentation, including random crop with padding 4 and random rotation between -10 and 10 degrees. The batch size is set as 64. For a visual comparison, the real images and the generated images by the SEC-CGAN are shown in Figure 3, indicating good quality of synthesized images for enhancing classifier training. The test results of all models are summarized in Table 1. The SEC-CGAN shows the best performance across the five dataset scenarios. When the training dataset is smaller, the SEC-CGAN performs much better than the other two models. For the 5% training dataset scenario, the SEC-CCAN shows 1.7% accuracy improvement over the EC-GAN, and 3.1% improvement over the baseline classifier. This indicates that the synthesized data generated by the SEC-CGAN is beneficial for training the classifier when the annotated data is limited.

**Table 1.** Classification accuracy with SVHN dataset.

| Training set | Baseline Classifier [32] | EC-GAN [11] | SEC-CGAN (proposed) |
|---|---|---|---|
| 5% | 87.2% | 88.6% | 90.3% |
| 10% | 90.4% | 91.6% | 91.8% |
| 15% | 91.9% | 93.2% | 93.6% |
| 20% | 93.0% | 93.4% | 94.0% |
| 25% | 93.2% | 93.4% | 94.2% |

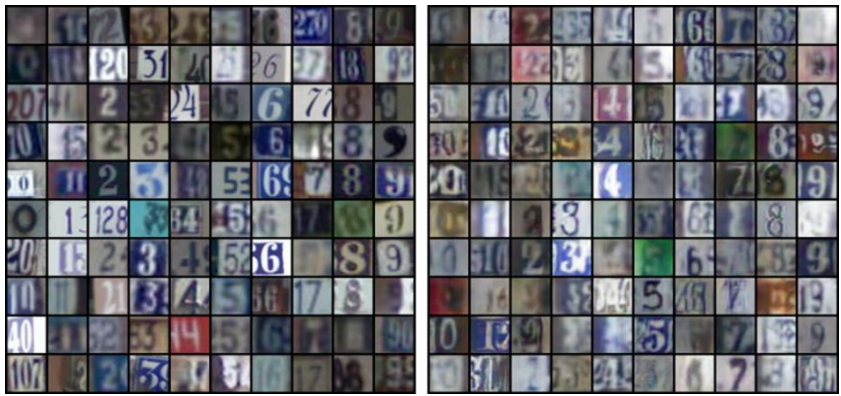

**Figure 3.** Left: images in the SVHN dataset; Right: synthesized images generated by SEC-CGAN; From left to right, images on each column denote class 0 to 9.

### 4.2. Highway image dataset

One of the motivations for this study is to classify different road weather conditions for travel alerts. It is well known that the accident risk increases substantially for people travelling under poor weather conditions, such as foggy, rainy, and snowy, when both the visibility and pavement surface friction are significantly impaired. Thus, accurate weather classification using live images from existing traffic cameras would be of practical importance for safety advisory. It helps to lower the accident risk by providing timely, location-specific risk warnings to the traveling public. For this reason, we trained the SEC-CGAN on a weather classification task with a self-collected highway image dataset, which was gathered from four states in the United States as summarized in Table 2. All the data were collected in the form of images from publicly accessible traffic cameras. The dataset includes four classes of weather types: foggy, rainy, snowy, and sunny. Examples of road weather images are shown in Figure 4.

**Table 2.** Sources of highway image dataset.

| Department of Transportation | Weather | Images |
| --- | --- | --- |
| California | foggy | 2,594 |
| Washington | rainy | 1,698 |
| Minnesota | snowy | 1,348 |
| Georgia | sunny | 3,901 |
| Total | | 9,541 |

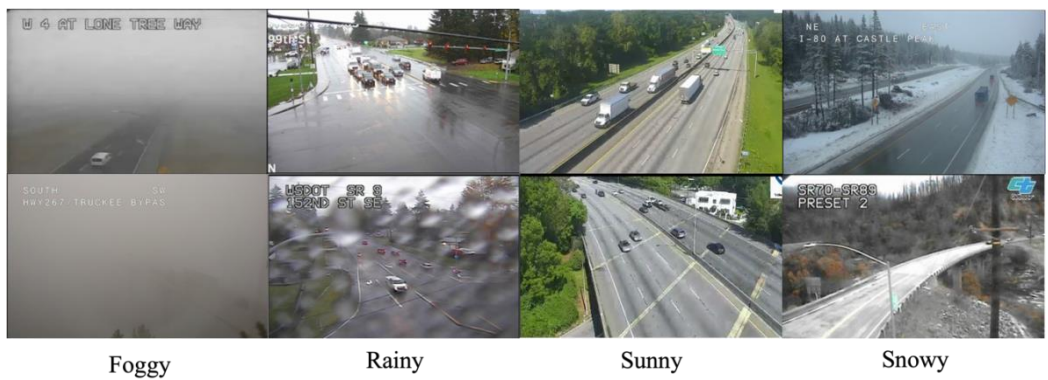

| Foggy | Rainy | Sunny | Snowy |

**Figure 4.** Sample images of road weather conditions.

For the training of the SEC-CGAN, the learning rate is set to 5e-5 for the discriminator and 2e-4 for the generator. In the parameters tuning trials, we found that a lower learning rate for the discriminator helps to prevent the collapse of conditional DCGAN. For data augmentation, a random horizon flip method is used. Since the collected images are in varied sizes, they are resized to 128 x 128 with a batch size of 64. Same as the training with SVHN dataset, the multiplier $\lambda$ is set to 0.6, and the threshold $\beta$ is set to 0.7.

Four train/test split scenarios are evaluated: 0.1/0.9, 0.2/0.8, 0.3/0.7, 0.4/0.6. The results are summarized in Table 3.

**Table 3.** Classification accuracy with highway image dataset.

| Dataset split (train/test) | Baseline Classifier [32] | EC-GAN [11] | SEC-CGAN (proposed) |
|---|---|---|---|
| 0.1/0.9 | 95.5% | 97.8% | 98.3% |
| 0.2/0.8 | 96.1% | 97.3% | 98.8% |
| 0.3/0.7 | 95.6% | 98.8% | 99.3% |
| 0.4/0.6 | 96.8% | 99.4% | 99.6% |

As shown in Table 3, with the 10% training dataset, the SEC-CGAN can reach 98.3% classification accuracy. The proposed SEC-CGAN outperforms the EC-GAN and baseline classifier in all four scenarios, evidencing the effectivity of the SEC-CGAN paradigm. The better performance of the SEC-CGAN and the EC-GAN over the baseline classifier attests to the benefit of using GAN-synthesized data in the classifier training.

For visualization purposes, exemplar synthesized images produced by the SEC-CGAN are shown in Figure 5. In each grid of 4 by 4, from left to right, they are synthesized images of foggy, rainy, snowy, and sunny weather conditions, respectively. It is clear that the synthesized images by the SEC-CGAN capture salient weather semantics that enhance classifier training and lead to improved classification performance.

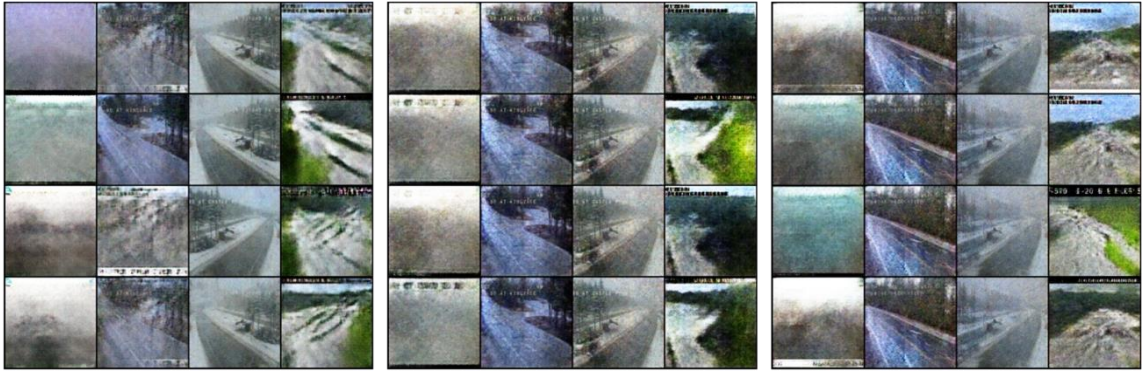

**Figure 5.** Three examples of synthesized images from the proposed SEC-CGAN. For each grid, the columns from left-to-right are: foggy, rainy, snowy, sunny.

## 5. Conclusions

In this paper, we introduce a SEC-CGAN method that employs a co-supervised paradigm and exploits the generative power of a CGAN to improve classification accuracy for small and imbalanced datasets. Particularly, the SEC-CGAN uses an independent classifier for the classification task while leveraging a conditional DCGAN to provide semantics-conditioned, confidence-aware synthesized examples. In this setting, the conditional DCGAN serves as a *co-supervisor* that controls which classes to generate and selects qualified synthesized images by their confidence scores. This leads to efficient sampling of minority classes and high quality of synthesized images for improving classifier training. Our experiments demonstrated that the SEC-CGAN is sample-efficient and outperforms EC-GAN, the state-of-the-art semi-supervised method for small-sample classification.

In summary, the practical benefit of the SEC-CGAN is two-fold: (1) generating distinct images to supplement small datasets along the classifier training process, in which a quality control for synthesized data is naturally embedded with the discriminator of conditional GAN (CGAN), and (2) creating more balanced class distribution through efficient sampling of minority classes during model training by delicately exploiting CGAN's inherent structure with labels. The former addresses the limitation of small datasets while the latter alleviates the issue of data imbalance, which is common in real-world settings.

In addition, the proposed SEC-CGAN paradigm is generic, end-to-end, flexible, and effective, and can be applied to various classification tasks across domains. For our experiments, the deep convolutional conditional GAN and ResNet-18 backbone are used in the proposed SEC-CGAN paradigm. Other network architectures can be used to suit task-specific applications for desirable performance.